\documentclass[10pt,twocolumn,letterpaper]{article}

\usepackage{iccv}
\usepackage{times}
\usepackage{epsfig}
\usepackage{graphicx}
\usepackage{amsmath}
\usepackage{amssymb}
\usepackage{threeparttable}
\usepackage{multirow}
\usepackage{pifont}
\newcommand{\cmark}{\ding{51}}%
\newcommand{\xmark}{\ding{55}}%
\usepackage{balance}

\usepackage[breaklinks=true,bookmarks=false]{hyperref}

\iccvfinalcopy 

\def\iccvPaperID{3128} 

\ificcvfinal\fi
\begin{document}

\title{Large-scale Tag-based Font Retrieval with Generative Feature Learning}



\author{
   Tianlang Chen$^{1}$\thanks{Work was done while Tianlang Chen was an Intern at Adobe.
} , Zhaowen Wang$^{2}$, Ning Xu$^{2}$, Hailin Jin$^{2}$ and Jiebo Luo$^{1}$\\
   $^1$University of Rochester 
   $^2$Adobe Research\\
   \small{\{tchen45,~jluo\}@cs.rochester.edu, \{zhawang,~nxu,~hljin\}@adobe.com}
}

\maketitle
\ificcvfinal\thispagestyle{empty}\fi

\begin{abstract}
Font selection is one of the most important steps in a design workflow. Traditional methods rely on ordered lists which require significant domain knowledge and are often difficult to use even for trained professionals. In this paper, we address the problem of large-scale tag-based font retrieval which aims to bring semantics to the font selection process and enable people without expert knowledge to use fonts effectively. We collect a large-scale font tagging dataset of high-quality professional fonts. The dataset contains nearly 20,000 fonts, 2,000 tags, and hundreds of thousands of font-tag relations. We propose a novel generative feature learning algorithm that leverages the unique characteristics of fonts. The key idea is that font images are synthetic and can therefore be controlled by the learning algorithm. We design an integrated rendering and learning process so that the visual feature from one image can be used to reconstruct another image with different text. The resulting feature captures important font design details while is robust to nuisance factors such as text. We propose a novel attention mechanism to re-weight the visual feature for joint visual-text modeling. We combine the feature and the attention mechanism in a novel recognition-retrieval model. Experimental results show that our method significantly outperforms the state-of-the-art for the important problem of large-scale tag-based font retrieval.

\end{abstract}

\section{Introduction}
\label{sec:intro}
\begin{figure}[!t]
\vspace{-2mm}
\centering
\includegraphics[width=3.3in]{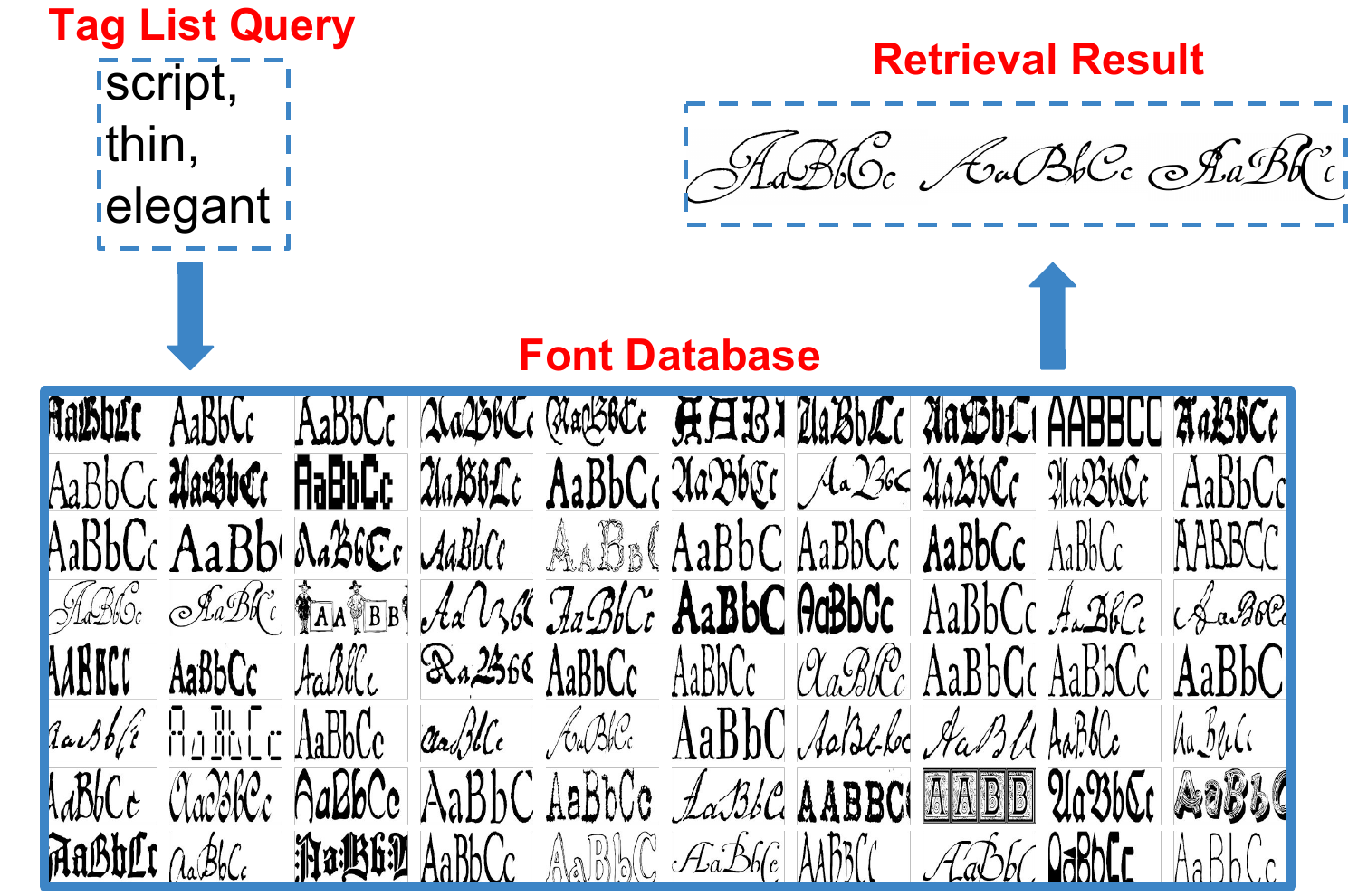}
\caption{Overview of large-scale tag-based font retrieval.}
\label{fig:intro}
\vspace{-5mm}
\end{figure}

Font is one of the most important elements in digital design. Designers
carefully choose fonts to convey design ideas. Since digital fonts were
invented in the 1950s, millions of fonts have been created to help
designers create engaging and effective designs. For instance,
MyFonts.com\footnote{www.myfonts.com}, one of the many font websites, offers over 130,000 fonts
with diverse designs, such as script, handwritten, decorative, just to
name a few categories. With the vast number of fonts available, font
selection becomes a challenge. Aspiring designers would take months to
learn typography and font selection as part of their design
curriculum. However, an average person who uses Microsoft Word will not
have time to learn typography but may have an intuitive idea about what
she wants to create. In this work, we study the problem of large-scale
tag-based font retrieval (illustrated in Figure~\ref{fig:intro}), which we
believe is an important step toward developing intuitive font selection
tools for average users.

Machine Learning approaches for tag-based font retrieval are first
studied by O'Donovan et al. in \cite{o2014exploratory} where they
collect a tag-based font dataset and use Gradient Boosted Regression
Trees \cite{friedman2001greedy}. However, this dataset only contains
1,278 fonts and 37 tags, which significantly limits its
applicability. In this work, we are interested in a broad range of tags
that a user can choose to express her design needs including properties,
functionalities, attributes, to subjective descriptions, emotions, etc.
To this end, we collect a large-scale font dataset\footnote{https://www.cs.rochester.edu/u/tchen45/font/font.html} from MyFonts.com,
which contains nearly 20,000 high-quality professional fonts and 2,000
unique user tags. This dataset will be released as a benchmark dataset
for tag-based font retrieval.

\begin{figure}[t!]
\centering
\includegraphics[width=3.2in]{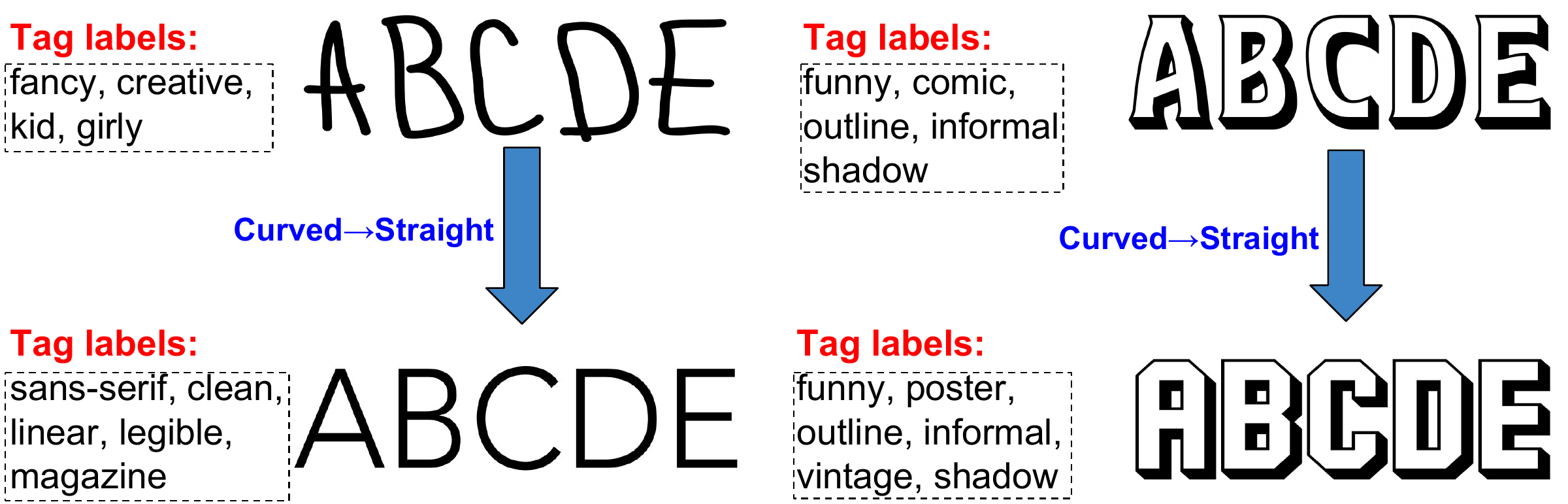}
\caption{An example to illustrate that a specific kind of visual feature
  can have different significance for tagging on different fonts.}
\vspace{-5mm}
\label{fig:problem}
\end{figure}

Feature plays a critical role in learning-based retrieval
algorithms. The state-of-the-art image retrieval methods use
convolutional neural networks to obtain image features which are
typically pre-trained on large-scale image classification datasets such
as ImageNet \cite{imagenet12}. One may attempt to follow this paradigm
for font retrieval. For instance, she can use the DeepFont feature
\cite{wang2015deepfont} which is trained to recognize fonts from text
images.  We find that although the DeepFont feature is learned to be
agnostic to characters, traces of characters remain in the feature.  For
instance, the features of one string with two different fonts is much
closer to the features of two different strings with the same font. Note
that this problem may not be unique in fonts but is certainly much more
severe in font retrieval than in image retrieval. The reason is that the
appearance variability of text images is much larger than that of object
images. In fact, there are an infinite number of text combinations.
Another major problem in feature learning for tag-based font retrieval
is that different parts of the font feature may have different
significance for different tags in terms of tag prediction. As shown in
Figure~\ref{fig:problem}, the change in line type from curved to
straight separates the upper two fonts from the bottom ones. At the same
time, the left two have very different tags, while the right two have
almost identical tags. This problem is usually addressed through the
visual attention mechanism \cite{you2016image, lu2017knowing,
  yang2016stacked, anderson2017bottom, lee2018stacked,
  li2017identity}. However, the challenge is to learn effective
attention models.

In this work, we propose a novel generative feature learning algorithm
for font retrieval that addresses the first problem. The
key idea is that font is synthetic and font images can be obtained
through a process that can be controlled by the learning algorithm. This
is a fundamental difference between real images and synthetic
images. Real images cannot be controlled by the learning algorithm.  We
design the learning algorithm and the rendering process so that the font
feature is agnostic to the characters.  Specifically, we take one font
and two characters and render three images. The first two images are
rendered with the given font and the two characters respectively.  The
third image is rendered with the same character of the first image and a
generic font. We train the network so that the feature extracted from
the first image together with that of the third image can generate the
second image using a combination of reconstruction and adversarial
losses. This way, the feature from the first image has to preserve the
font information but not the character information because it has to
reconstruct the second image which has a different character. To
address the second attention problem, we observe that there exists a
strong correlation between font recognition and attention maps.  Based
on this observation, we design an implicit attention mechanism to help
the model adaptively select useful information from the font recognition
feature. In particular, we compute an attention map re-weights the
feature as an attentive selection process.

\subsection{Main Contributions}
The main contributions of our paper are as follows: 1.  We collect a
large-scale font tagging dataset of high-quality professional fonts. The
dataset contains nearly 20,000 fonts and 2,000 tags and a human curated
evaluation set. 2. We propose a generative feature learning
algorithm that leverages the unique property of font images.  3. We
propose an attention mechanism to re-weight the learned feature for
visual-text joint modeling. 4. We combine the two components in a
recognition-retrieval model that achieves top performance for font
retrieval.

\section{Related Work}
\label{sec:related}

The combination of font understanding and machine learning starts from
font recognition, where the problem is to recognize a font from a text
image. Early font recognition works try to recognize a font via
artificial font features. Zramdini et al.~\cite{zramdini1998optical}
identify the typeface, weight, slope, and size of a font image and use a
multivariate Bayesian classifier to classify the font. Zhu et
al.~\cite{zhu2001font} leverage Gabor Filters to extract
content-independent features for recognition. Chen et
al.~\cite{chen2014large} feed local feature metric learning into a
nearest class mean font classifier. As deep learning becomes popular,
Wang et al. \cite{wang2015deepfont} build a Convolution Neural Network
with domain adaptation techniques for font recognition, while Liu et
al. \cite{liu2018font} use a multi-task adversarial network to learn a
disentangled representation and apply it to recognize Japanese fonts. In \cite{Liu_2018_ECCV}, Liu et al. leverage GAN to perform one-way transform from scene text to clean font image for better recognition. Compared with \cite{Liu_2018_ECCV}, we essentially perform font-specific mutual glyph transform as a bridge to connect the glyph-level input and the font-level tag.
On the other hand, font generation and font style transfer has became
hot spot topics in recent years. The work of Azadi et
al.~\cite{azadi2018multi} successfully generates unobserved glyphs from
restricted observed ones through an stacked conditional GAN model, Zhang
et al. \cite{zhang2018separating} propose an encoder-decoder EMD model
which separates content and style factors for font style
transfer. Following their work, we focus on tag-based font retrieval
introduced in \cite{o2014exploratory} and propose a deep learning
approach. Compared with other attribute-based retrieval problems such as
bird \cite{WahCUB_200_2011}, scene \cite{zhou2018places}, and animal
\cite{lampert2009learning}, there are larger numbers of tags organized
in complex semantic relationship, which significantly increases the
difficulty of this problem. While most recent work
\cite{zhang2017learning,socher2013zero,xian2017zero} on the mentioned
datasets focus on zero-shot image retrieval , we currently do not
consider unseen tags for this task since there is no specialized
knowledge base to describe font-related tags. But it is a problem of
future interests.


Generative Adversarial Network (GAN) is porposed by Goodfellow et
al. \cite{goodfellow2014generative}, which makes a generative model and
a discriminative model play a minimax game to encourage the generative
model to output desired synthetic images. Recently, various GAN
structures are proposed that successfully achieve paired and unpaired
image-to-image transformation \cite{isola2017image, zhu2017unpaired},
and image generation conditioned on class labels
\cite{mirza2014conditional}. The idea of GAN and adversarial learning
are also applied on image retrieval task. Wang et
al. \cite{wang2017adversarial} propose an ACMR method for cross modal
retrieval, which implements adversarial learning to reduce the domain
gap between the text feature and the image feature so that a shared embedding
space is constructed. Gu et al. \cite{gu2018look} achieve similar goal
by directly integrating a GAN to generate corresponding images from text
feature. The work of Zhang et al. \cite{zhang2018attention} train an
attention module and a hashing module in an adversarial way, which
guides the attention module to focus on useful regions/words of an
image/text.

Attention mechanism has been successfully applied to different
visual-textual joint learning tasks such as image captioning
\cite{you2016image,lu2017knowing}, visual question answering
\cite{yang2016stacked,anderson2017bottom}, text-based image retrieval
\cite{lee2018stacked,li2017identity} and semantic
synthesis~\cite{ma2018gan}. In most situations, attention mechanism
directly guides the model to capture useful image regions or language
segments, which can be visualized and considered as explicit
attention. On the other hand, Kim et al.~\cite{kim2016multimodal}
propose an implicit attention mechanism, which generates the attention
weight on feature maps instead of on raw data. Inspired by this work, we
design an effective approach to select useful features adaptively for
each input font image.

\section{Large-scale Font Tagging Dataset}
\label{sec:dataset}

To the best of our knowledge, there exists no previous public
large-scale dataset suitable for tag-based font retrieval. The only
font-tag dataset available is the one collected by O'Donovan et
al. \cite{o2014exploratory}. It contains 1,278 Google web fonts and 37
attributes. To facilitate research in this direction, we collect a
benchmark font tagging dataset from
\href{http://www.myfonts.com}{MyFonts.com}. We select MyFonts for the
following reasons: (1) Compared with free font resources such as Google
fonts, the fonts on MyFonts are designed by well-known commercial font
foundries which are in the typography business for decades and use by
professional graphic designers. (2) MyFonts allows foundries, designers,
and users to label fonts using arbitrary tags instead of selecting tags
from a pre-defined set. This tremendously expands the tag vocabulary of
the font retrieval system and enables semantic search of fonts. The
fonts on MyFonts are originally labeled according to font
families. Specifically, each font family contains a set of fonts of
different design variations (e.g. regular, italic, and bold). A family
is labeled by a list of associated tags. We find that under most
conditions, a font family is labeled based on the regular version of the
family. Therefore, for each font family, we use a set of pre-defined
rules to find the regular version and render text images of the regular
font as visual samples. In the end, we obtain in total 18,815 fonts and their
corresponding labeled tag lists. We collect the complete Roman character
glyph set (including both uppercase and lowercase of the 26 characters)
for each font and render images of random character strings for all the
fonts.  We randomly split the entire dataset into training, validation,
and test sets according to the proportion of 0.8/0.1/0.1.

\begin{figure}[!t]
\centering
\hspace{-1mm}\includegraphics[width=3.3in]{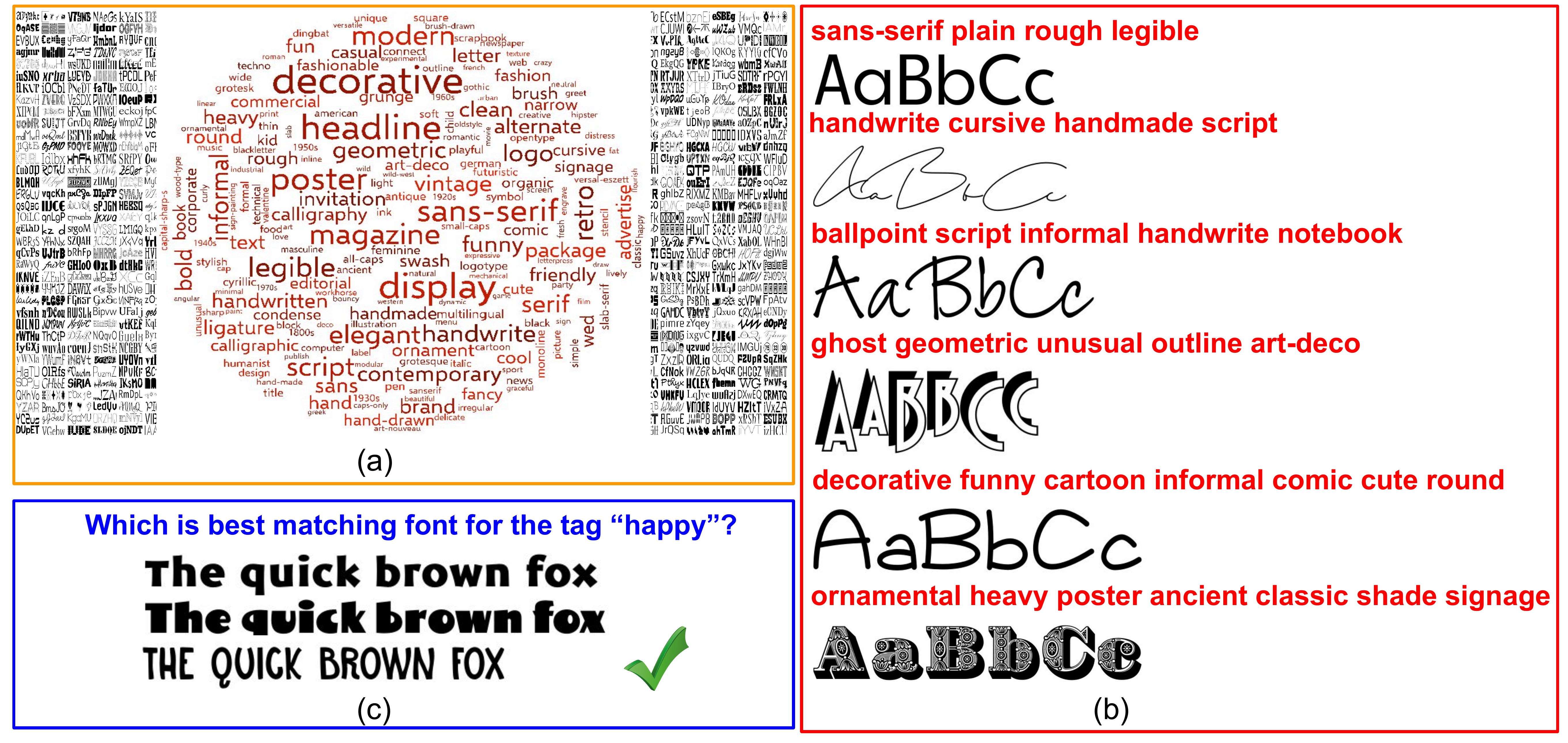}
\caption{Overview of our large-scale font tagging dataset: (a) A word
  cloud of high-frequency tags; (b) Font samples with labeled tags; (c)
  A sample font group for AMT workers to rank.}
\label{fig:dataset}
\vspace{-4mm}
\end{figure}

As each font may be tagged by different foundries, designers, and users with varying quality, to guarantee the tag consistency and filter out the
noisy labels, we sequentially apply the following tag pre-processing
procedures: (1) convert all the words to lowercase and correct
misspelling; (2) lemmatizing every word (e.g. kids $\rightarrow$ kid);
(3) converting a N-Gram tag into a word with hyphens (e.g. sans serif
$\rightarrow$ sans-serif); (4) combining the same tags based on the
first three steps; and (5) removing infrequent tags occurring less than
10 times in the training set. In the end, we obtain 1,824 tags for our
dataset. An overview of our collected dataset is shown as
Figure~\ref{fig:dataset}.

It should be noted that the tag data from MyFonts only indicate whether
a font matches a tag, but do not include the relative ranking of the
matching. Moreover, tag labels collected from the web are inevitably
noisy and incomplete. For the purpose of evaluating algorithms, we
additionally collect a high-quality tagging set via Amazon Mechanical
Turk (AMT). This additional tagging dataset contains the
ranking information of different fonts for a given tag. Specifically, we
choose the top-300 frequent tags of the MyFonts dataset. For each tag,
we randomly generate 30 groups, each of which contains three test fonts
that are already labeled with this tag. For each group of fonts, three
AMT workers are requested to select the best matching font to the given
tag after viewing a list of training fonts labeled with this tag as a
reference. To avoid bias from text semantics, each font is represented
by a font image rendered with a standard text as in
Figure~\ref{fig:dataset}. We add a group into the test set and label one
of the fonts as the ground-truth only when all three workers select
this font from this group. In the end, we collect 1,661 high confidence
groups for the 300 tags, which we use as the evaluation test set in the
experiments.

\section{Our Method}


\begin{figure}[!t]
\vspace{-3mm}
\centering
\includegraphics[width=3.3in]{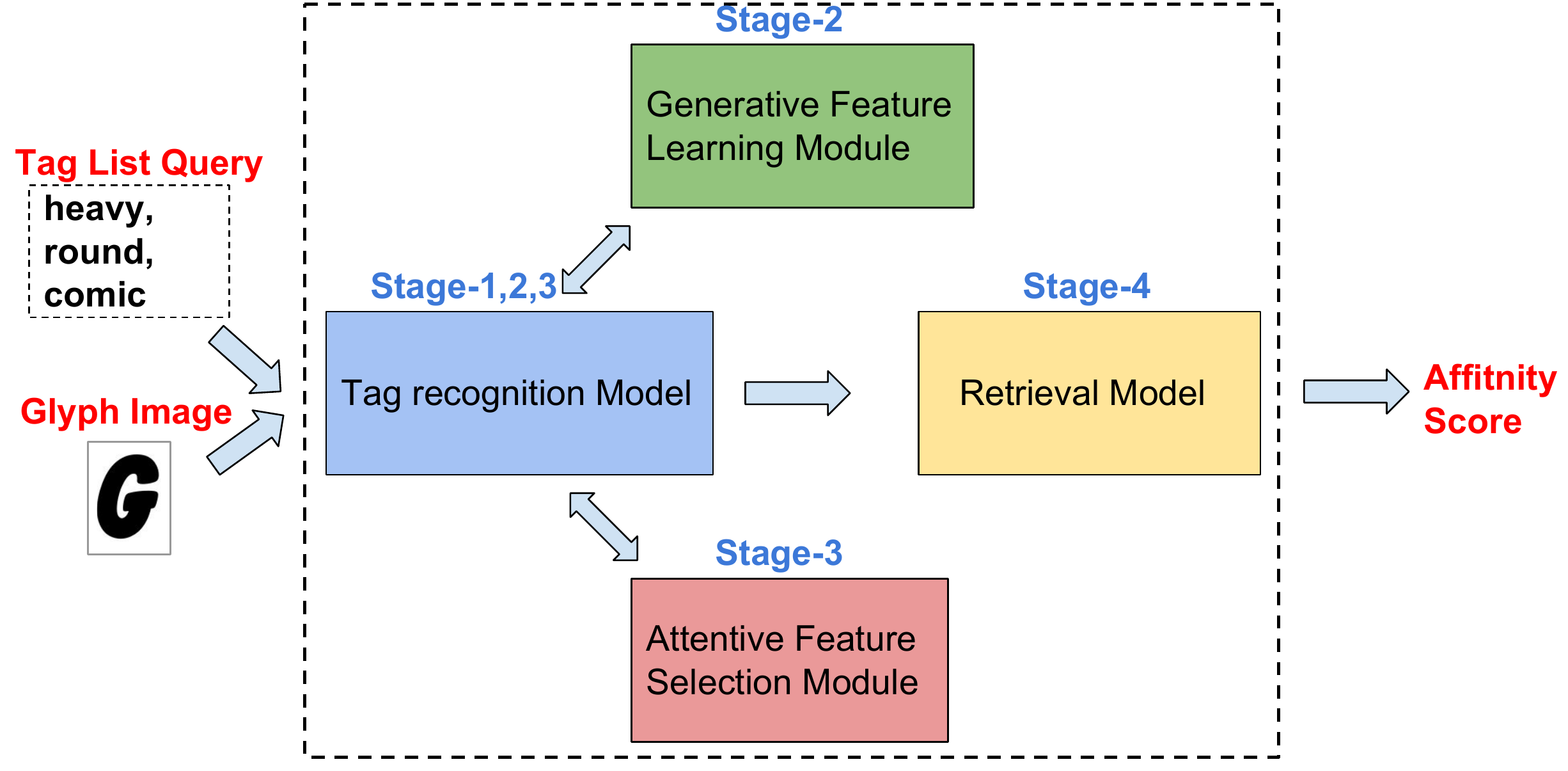}
\caption{Overview of the proposed font retrieval system. We illustrate the stages where a specific module is trained. }
\label{fig:overview}
\vspace{-3mm}
\end{figure}

In this paper, we follow the common setting of image retrieval tasks \cite{yang2018learning,li2017identity,wang2017adversarial,li2017person}. We train a model to predict an affinity score between an input query and a a font as their similarity. The font is represented by 52 glyph images (a-b and A-B). The affinity score between a font and a query is averaged over all the 52 glyph images belonging to this font. When a user inputs a query, we compute the affinity scores of this query with all the fonts in the system and recommend the ones with the highest affinity scores. In contrast to the previous work ~\cite{wang2015deepfont} where each image contains multiple characters, we find the global structure of a text image (its character composition) does not provide extra information for the font retrieval problem. The overall architecture of the proposed model is illustrated in Figure~\ref{fig:overview}, which is sequentially trained in four stages. In the following subsections we describe each stage in details.

\subsection{Basic Tag Recognition Model} \label{sec:rec}
In the first stage, different from typical retrieval models \cite{reed2016learning, wang2017adversarial,zheng2017dual} that directly learn a joint embedding for image and tag features, we train a tag recognition model which predicts the probability of an input glyph image with respect to each tag. Overall, this is a multi-label learning task, since one image may correspond to multiple tags. Specifically, 
let $\{F_{1}, ..., F_{M}\}$ be the training font set and $\{L_{1}, ..., L_{52}\}$ be the glyph set. We denote the text image containing glyph $L_{j}$ of font $F_{i}$ as $I_{i}^{j}$. Our basic tag recognition model first extracts a hidden feature $f_{i}^{j}$ using a convolutional neural network. The feature is then fed into a fully-connected layer with $N$ output nodes, where $N$ is the total tag vocabulary size. In the end, a sigmoid unit maps the value of each node to the range of $(0, 1)$, which corresponds to the tag probability for the input image. We employ a cross-entropy loss to train the model:
\begin{align} 
L_{c}=\sum_{i,j}\sum_{k=1}^{N}(y_{i}^{k}\log(p_{i}^{j,k})+(1{-}y_i^k)\log(1{-}p_i^{j,k})), \label{equ:ce}
\end{align}
where $p_{i}^{j,k}$ is the predicted probability for $I_{i}^{j}$ wrt the $k^{th}$ tag; $y_{i}^{k}$ is 1 if $F_{i}$ is labeled with the $k^{th}$ tag, and 0 otherwise.


\subsection{Generative Feature Learning}
\label{sec:collect}

\begin{figure}[!t]
\centering
\includegraphics[width=3.3in]{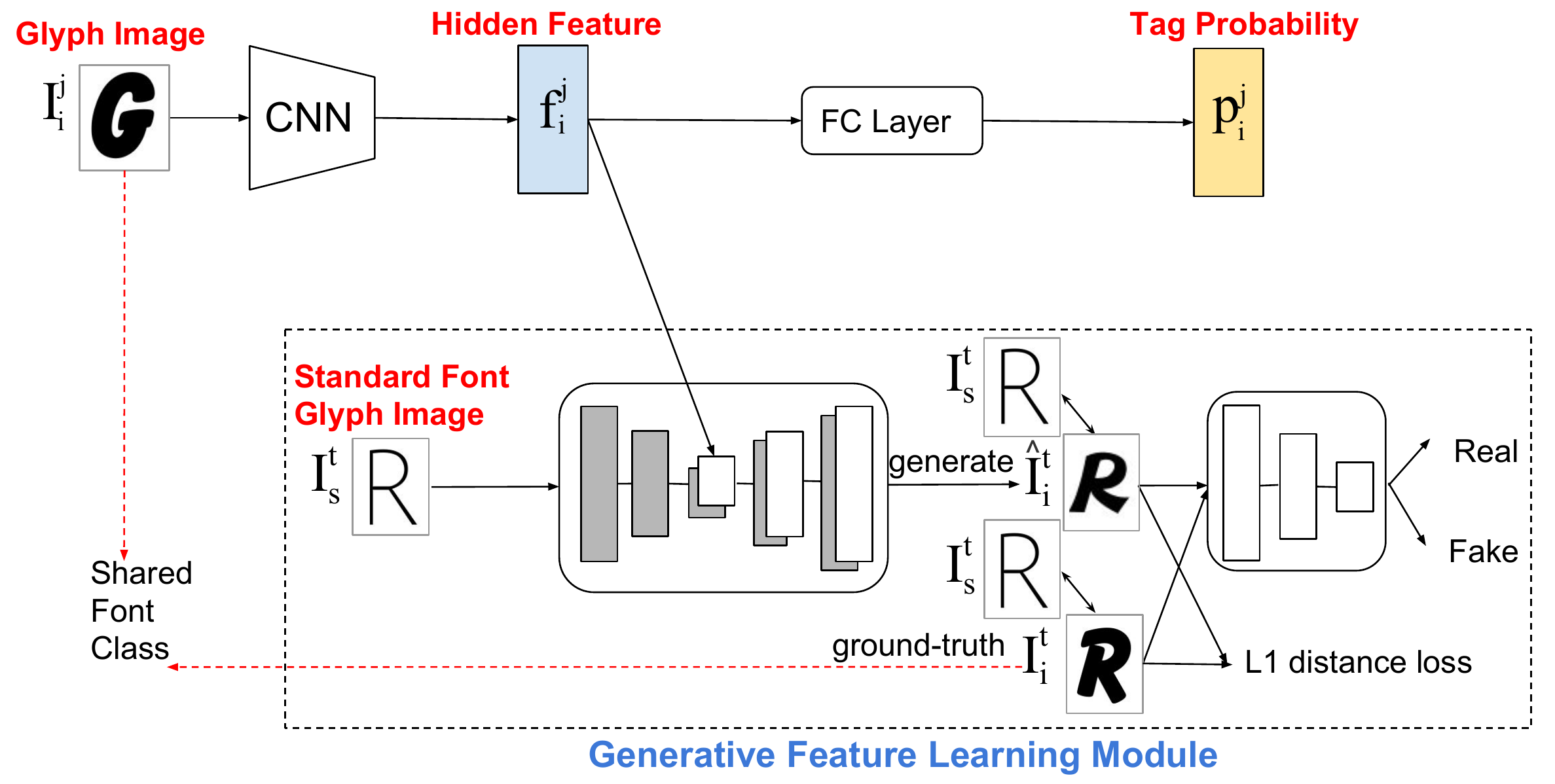}
\caption{Generative Feature Learning: we transform input glyph image's character with a generative adversarial network.}
\label{fig:gene}
\vspace{-3mm}
\end{figure}

After training the tag recognition model, in the second stage, we further encourage the extracted latent feature of a glyph image to better capture the information about the font  instead of the character content. Otherwise, it may bias tag prediction with irrelevant information. Our idea is that we require a glyph image to be accurately generated via the latent feature of another glyph image with a different character but the same font. In particular, we design a conditional generative adversarial network \cite{goodfellow2014generative} consisting of a generator and a discriminator as in Figure~\ref{fig:gene}. Conditioned on the feature $f_i^j$ of image $I_i^j$, the generator $G$ is trained to convert an input image $I_s^t$ rendered with a fixed standard font $F_s$ and an arbitrary character $L_t$ to a target image $\hat{I}_i^t$ of the same font style as $I_i^j$ and same character as $I_s^t$ (i.e. $\hat{I}_i^t$ = G($f_i^j$, $I_s^t$)). The discriminator $D$ is trained to discriminate between the generated image $\hat{I}_i^t$ and the real image $I_i^t$. The objective of the GAN is expressed as:
\begin{small}
\vspace{-1mm}
\begin{align}
L_{gan}=\min_{G}\max_{D}(&\mathbb{E}_{I_{i}^{t}, I_{s}^{t}}[\log{D(I_{i}^{t}, I_{s}^{t})}] +\\&\mathbb{E}_{f_{i}^{j}, I_{s}^{t}}[\log{(1-D(G(f_{i}^{j}, I_{s}^{t}), I_{s}^{t}))}]) \notag
\end{align}
\end{small}
Following \cite{isola2017image}, we design a PatchGAN architecture for the discriminator to enhance the local details of the generated image. Random noise is provided in the form of dropout. Compared with DR-GAN \cite{tran2017disentangled}, we do not require the discriminator to explicitly recognize the character class of the generated glyph image, which is irrelevant in our case. To further encourage the generated image to have the same font style as the targeted one, we add an $L_1$ distance loss as \cite{isola2017image}:
\begin{align}
L_{L_1} = \mathbb{E}_{I_{i}^{t}, \hat{I}_{i}^{t}} \big(\| I_{i}^{t} - \hat{I}_{i}^{t} \|_{1} \big).
\end{align}     
\vspace{-1mm}
Our final objective of the GAN branch is:
\vspace{-1mm}
\begin{align}
L_{lgan} = L_{gan} + \lambda \cdot L_{L_1} \label{eqa:gan}
\end{align}
\vspace{-1mm}
The training objective of the recognition model is:
\vspace{-1mm}
\begin{align}
L_{rec} = L_{c} + \beta \cdot L_{lgan}, \label{eqa:collect}
\end{align}
where $\lambda$ and $\beta$ are hyper-parameters to adjust the relative weights among different loss terms. In this training stage, we first train GAN using Equation~\ref{eqa:gan}, and the features $f_i^j$ are taken from the pre-trained tag recognition model without back-propagation. After convergence, the tag recognition model is fine-tuned by the ground-truth tags and GAN alternately. More concretely, we split each training epoch into two sub-epochs. In the first sub-epoch, we optimize the parameters of the CNN by $\beta \cdot L_{lgan}$ while fixing the last fully-connected layer. In the second sub-epoch, we optimize the parameters of the CNN and the fully-connected layer by $L_{c}$. This training strategy yields better performance than a multi-task training that jointly updates all the parameters.

\subsection{Attentive Feature Selection }\label{sec:select}

\begin{figure}[!t]
\centering
\includegraphics[width=3.4in]{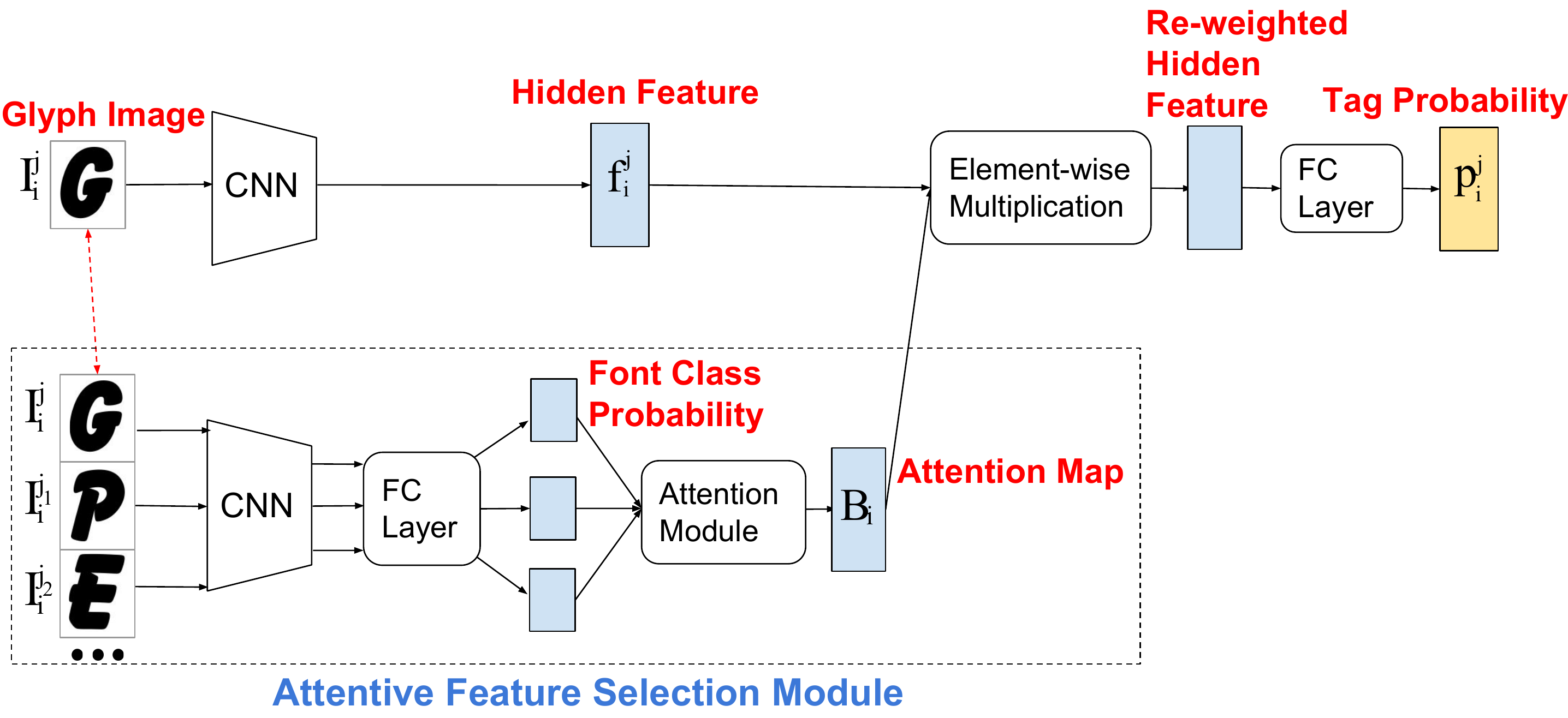}
\vspace{-1mm}
\caption{Attentive Feature Selection: an implicit attention mechanism adaptively selects useful features of the glyph image.}
\vspace{-3mm}
\label{fig:att}
\end{figure}

In the third stage, as motivated by the problem shown in Figure~\ref{fig:problem}, our model is further guided to select relevant information based on the font class prediction of the input image. To this end, we train a font classification model to predict the font type of the input glyph image as in Figure~\ref{fig:att}. The structure of the font classification model is similar to that of the tag recognition model. It integrates a CNN to extract the hidden feature of the image and feeds it into a fully-connected layer with an output node size equal to the total number of font classes. Since one glyph image uniquely belongs to one font, this task is a single-label classification task. We thus replace the sigmoid unit with the softmax unit to map the outputs of the fully-connected layer, and train the font classification model using the cross-entropy loss. We feed the predicted font class probability distribution $c_i^j$ for an input glyph $I_i^j$ into an attention module which contains a fully-connected layer followed by a sigmoid unit. The distribution is transformed into an attention map $B_i^j$ where $B_i^j$ has the same dimension as the hidden feature $f_i^j$, and its values are in the range of $(0,1)$.

The attention map estimated based on one image $I_i^j$ may not be reliable. Therefore, during training, we aggregate the attention maps from a set of $J$ images $\{I_i^{j_1}, ..., I_i^{j_J}\}$ of the same font  $F_i$ and randomly selected character set $L_{j_1}, ..., L_{j_J}$. The final attention map $B_i$ for font $F_i$ is computed as:
\begin{small}
\begin{align}
B_{i} = B_{i}^{j_1} \odot  B_{i}^{j_2} \odot ... \odot B_{i}^{j_J},
\end{align}
\end{small}
where $\odot$ represents the element-wise multiplication. This random selection of multiple glyph images improves the accuracy and selectivity of the attention map for a specific font. In the end, we take the element-wise multiplication between $f_i^j$ and $B_i$ to obtain the re-weighted feature of $I_i^j$, which is then fed into the top fully-connected layer of the tag recognition model. As in \cite{kim2016multimodal}, our node-level feature re-weighting enforces an implicit attention mechanism. Figure~\ref{fig:att} shows the overall structure of our tag prediction model with the attentive feature selection module integrated. When we train the model in this stage, the same training objective as Section~\ref{sec:rec} is employed to update only the parameters of the attention module and the last tag recognition fully-connected layer. Given a glyph image at test time, we only extract one attention map from this single image without further aggregation, leading to a significantly faster retrieval speed as well as competitive accuracy.

\subsection{Combined Recognition-Retrieval Model}
\label{sec:ret}
For a single-tag query, we define its affinity score to a glyph image as the predicted probability of the tag recognition model. Indeed, the model can also be used for retrieving fonts from multi-tag queries by computing the affinity score as the product or sum of the predicted probabilities of all the tags in the query. However, due to the imbalance of tag occurrence, the predicted probabilities of popular tags are usually much higher than those unpopular ones, leading to the tag dominance problem. The top recommended fonts may just match few tags of a multi-tag query.



\begin{figure}[!t]
\centering
\includegraphics[width=3.3in]{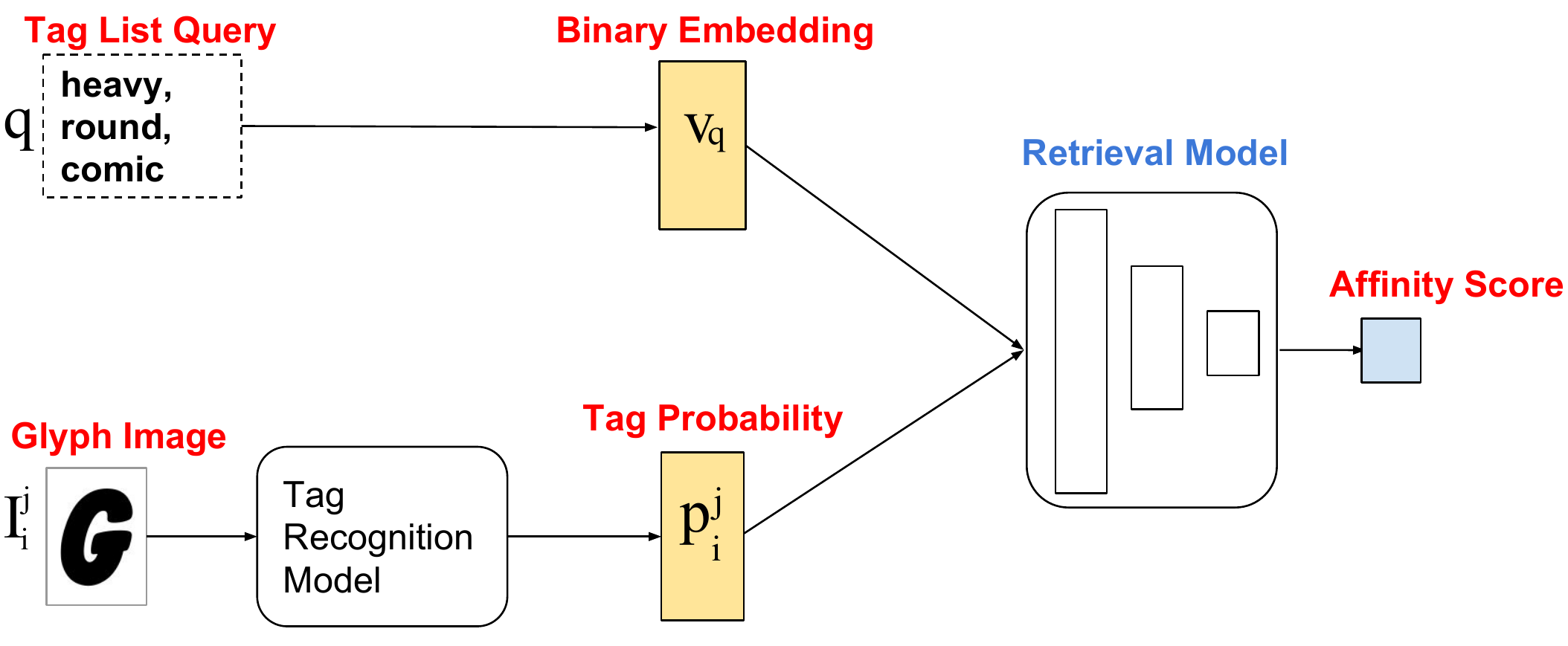}
\vspace{-1mm}
\caption{Combined recognition-retrieval model: the retrieval model is trained on top of the tag recognition model with query tag combinations.}
\vspace{-3mm}
\label{fig:ret}
\end{figure}

Therefore, for multi-tag query, in the fourth stage, we introduce a retrieval model (as Figure~\ref{fig:ret}) on top of the tag recognition model that maps the predicted tag probabilities to a comprehensive affinity score with unbiased consideration for each query tag. Given a pair of input image $I_i^j$ and query $q$, the tag recognition model first predicts the tag probability distribution $p_i^j$ $\in$ $\mathbb{R}^N$. On the other hand, the query is encoded as a binary vector $v_{q}$ $\in$ $\mathbb{R}^N$, whose $t$-th entry is set to 1 if the query contains the $t$-th tag. The retrieval model takes the element-wise multiplication of $p_i^j$ and $v_q$ to generate a query-based tag probability vector, and feeds it into two-layer fully-connected network to get the affinity score between $I_{i}^{j}$ and $q$. Before fed into the fully-connected layer, the query-based tag probability vector is first mapped by a power activation function $x\rightarrow (x+\epsilon)^{\alpha}$, where $\alpha$ is a hyper-parameter and $\epsilon$ is used to prevent infinite gradient. It shows better performance in transforming probabilities than other common activation functions. At the top of the second layer there has a sigmoid activation function which maps the output score in the range of $(0, 1)$. Following \cite{yang2018learning}, we train the retrieval model as a relation network by sampling triplets of query, positive image and negative image. The query is composed of 2 to 5 randomly selected tags, all/not all of which are included in the ground-truth tags of the positive/negative image. We train the retrieval model by minimizing the following pairwise soft ranking loss:
\begin{align}
L_{ret} = \mathbb{E}_{q, I^+, I^-} \left[\log(1+\exp(\gamma(s(q, I^-){-}s(q, I^+)))\right] ,
\end{align}
where $I^+$ and $I^-$ are positive and negative samples for query $q$, and $s(q, I)$ is the affinity score between $q$ and $I$ outputted by the retrieval model. The parameters of the tag recognition model are fixed in this stage.

\section{Experiments} \label{sec:exp}

In this section, we compare our proposed method with state-of-the-art image retrieval methods adapted to our font retrieval dataset. An ablation study is presented to demonstrate the effectiveness of each component in our method. 

\subsection{Dataset and Experiment Settings}

We train all our models on the MyFonts training set and evaluate them on the two test sets presented in Section~\ref{sec:dataset}, which are named MyFonts-test and AMT-test, respectively.
MyFonts-test contains 1,877 fonts split from the original MyFonts dataset and is labeled with 1,684 different tags. We construct three query sets for evaluation. The first query set focuses on single-tag queries. It contains all the 1,684 tags with each tag considered as a query. The second query set focuses on multi-tag queries. For each font in MyFonts-test, we randomly generate 3 subsets of its ground-truth tags as 3 multi-tag queries. The query size ranges from 2 to 5 tags. After filter out repeating queries, this query set contains 5,466 multi-tag queries. The third query set focuses on testing the retrieval performance of the model on frequent tags that are more likely be searched. It contains the top-300 frequent tags in the training set with each tag considered as a query. For the first and third query sets, a font matches to a query (i.e. a positive font) if its ground-truth tag list contains the corresponding tag. For the second query set, a font matches to a query if its ground-truth tag list contains all the tags of the query. For each query set, we adopt two well-known retrieval metrics, \ie, Mean Average Precision (mAP) and Normalized Discounted cumulative gain (nDCG) for model evaluation. Please refer to our supplementary material for a detailed description of the two metrics.
AMT-test contains 1,661 groups, with each group associating a tag with one most relevant ground-truth font. Given a tag from a group, we compute and rank the affinity scores of all the fonts with respect to the tag for different models. We compare both the accuracy of each model selecting the ground-truth font in each group and the average rank of the ground-truth. 

\subsection{Implementation Details}

For both tag recognition model and font classification model, we select ResNet-50 \cite{he2016deep} as the base CNN architecture, where the hidden feature of an image is extracted from the pool5 layer of ResNet-50. We follow the pix2pix GAN \cite{isola2017image} to build the generator as an encoder-decoder structure with skip connections. It receives a standard 128 $\times$ 128 glyph image with a fixed font $F_{s}$ from its encoder, and concatenates the hidden feature computed by the tag recognition model before the bottleneck layer. For the discriminator, we apply a 14 $\times$ 14 local discriminator with three convolutional layers, to differentiate between real and fake local patches. We use hyper-parameters $\lambda{=}10$ in Equation~\ref{eqa:gan}, $\beta{=}0.04$ in Equation~\ref{eqa:collect}, $J{=}4$ in Section~\ref{sec:select}, and $\alpha{=}0.1$, $\gamma{=}100$ in Section~\ref{sec:ret}.
When being trained in a particular stage, the learning rates of the CNN, the GAN, the last fully-connected layer, and the top retrieval model are set to $5{\times}10^{-4}$, $2{\times}10^{-3}$, $5{\times}10^{-3}$ and $5{\times}10^{-3}$. We use batch size 20 in all the training experiments. Weight initialization details are provided in the supplementary material.

\subsection{MyFonts Test}

\begin{table}[t!]
  \centering\small
  \hspace{-2mm}
  \begin{threeparttable}
  \scalebox{0.82}{
  \begin{tabular}{|c|c|c|c|c|c|c|c|c|}
  \hline
    \cline{1-9}
    \multicolumn{3}{|c|}{Methods} &\multicolumn{2}{|c|}{Single tag (300)} &\multicolumn{2}{|c|}{Single tag (full)} &\multicolumn{2}{|c|}{Multi tag} \\\cline{4-9}
    \multicolumn{3}{|c|}{}&mAP&NDCG&mAP&NDCG&mAP&NDCG \\ \cline{1-9}
    \multicolumn{3}{|c|}{GNA-RNN~\cite{li2017person}}&14.90&56.97&5.03&28.41&7.01&27.49\\ \cline{1-9}
    \multicolumn{3}{|c|}{DeViSE~\cite{frome2013devise}}&13.43&55.98&3.73&26.04&5.48&25.80\\ \cline{1-9}
    \multicolumn{3}{|c|}{RelationNet~\cite{yang2018learning}}&15.33&57.49&5.66&29.27&7.52&28.05\\ \cline{1-9}
    \multicolumn{3}{|c|}{\textbf{Ours}}& \multicolumn{6}{|l|}{}\\ \cline{1-3}
    {GAN} & {Att} & {Retr} & \multicolumn{6}{|l|}{} \\ \cline{1-9}
    
    \xmark&\xmark&\xmark &26.29&68.67&16.77&42.63&14.93&35.52\\ \cline{1-9}
    \cmark&\xmark&\xmark &26.99&69.18&17.30&43.15&15.37&35.87\\ \cline{1-9}
    \xmark&\cmark&\xmark&27.75&69.81&17.76&43.65&15.78&36.40\\ \cline{1-9}
    \cmark&\cmark&\xmark&\textbf{28.08}&\textbf{70.04}&\textbf{18.02}&\textbf{43.95}&16.06&36.72\\ \cline{1-9}
 \cmark&\cmark&\cmark&\textbf{28.08}&\textbf{70.04}&\textbf{18.02}&\textbf{43.95}&\textbf{16.74}&\textbf{37.57}\\ \cline{1-9}
    \hline
    \end{tabular}
    }
  \end{threeparttable}
  \vspace{1mm}
  \caption{\label{tab:result-1} Comparison of different retrieval models on MyFonts-test set. ``GAN" denotes our generative feature learning module. ``Att" denotes our attentive feature selection module. ``Retr" denotes our retrieval model. ``\cmark" denotes with while ``\xmark" denotes without. For a framework without the retrieval model, it computes the affinity score between a glyph image and a query as the product of the image's predicted probabilities of all the tags in the query. }
  \vspace{-2mm}
\end{table}

\begin{figure*}[t!]
\centering\hspace{-3.5mm}
\includegraphics[width=7.0in]{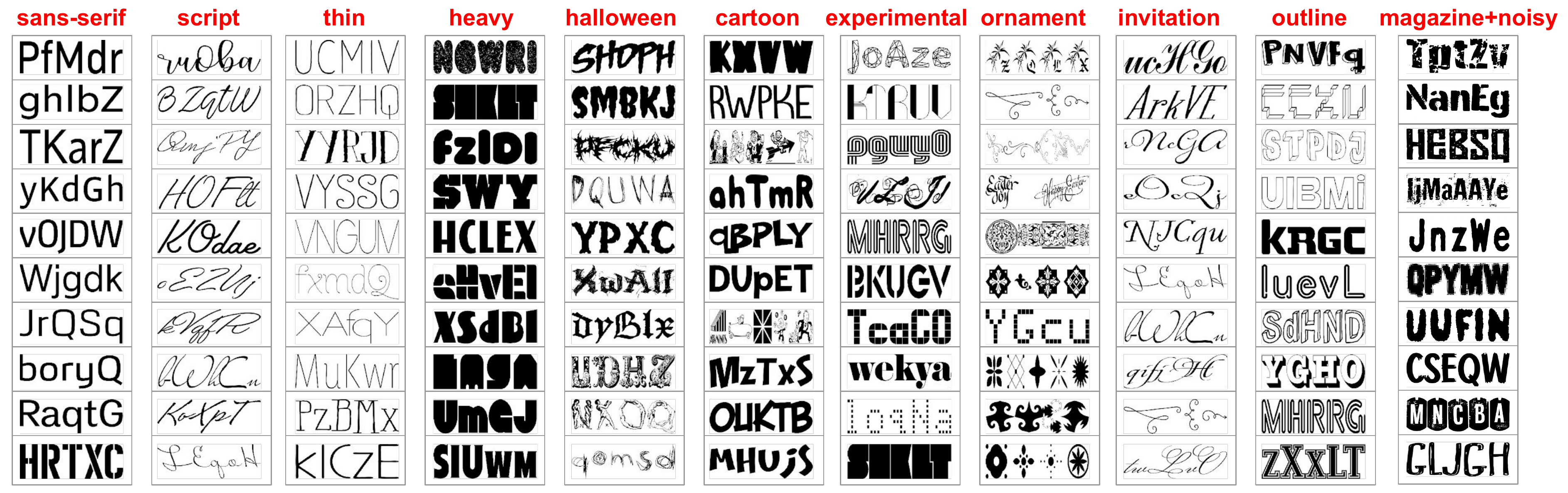}
\vspace{-4mm}
\caption{ Retrieval results of our model on typical single-tag and multi-tag queries (font is represented by a random 5-character image).  }

\label{fig:example}
\vspace{-4mm}
\end{figure*}

Since there are no prior works for large-scale font retrieval, we adapt and compare against several state-of-the-art image retrieval methods~\cite{li2017person,frome2013devise,yang2018learning} for our task. The implementation details are as follows.

\textbf{GNA-RNN} \cite{li2017person}: the modified GNA-RNN model uses ResNet-50~\cite{he2016deep} to extract the font image feature. Because there is no order among query tags, we use two fully-connected layers to extract the query feature from the binary encoding of the query input instead of using a RNN. After that, the image feature and query feature are mapped into the same embedding. Their inner product is further computed to obtain the affinity score. We use random positive and negative pairs of font images and query tags as training samples to train the model using a cross-entropy loss.

\textbf{DeViSE} \cite{frome2013devise}: Image features are extracted using the same ResNet-50 model. Tag embedding is trained from scratch because there is no appropriate text dataset with font related words. The method directly transforms an image and a query feature into an affinity score using a transformation matrix. Hinge ranking loss is applied during training. 

\textbf{RelationNet} \cite{yang2018learning}: We use the same networks as in {GNA-RNN} to extract and map the font features and the tag features. The two features are concatenated and mapped to affinity scores using two fully-connected layers. Mean square error loss is employed during training. 

The results of our method compared with other retrieval methods on the MyFonts-test set are presented in Table~\ref{tab:result-1}. It is clear that our method outperforms all the other methods by a large margin. Surprisingly, our basic tag recognition model described in Section~\ref{sec:rec} (the 4th result in Table~\ref{tab:result-1}) is already better than the other image retrieval based methods. We believe that the recognition based model is more suitable for the font retrieval task than the retrieval models which usually learn a joint embedding between the image feature and the tag feature. This can be attributed to the fact that font tags have very subtle meaning which is hard to capture by a joint embedding. In addition, the training data we collected from the web may not be complete. For example, some tags may be omitted by web users for a given font. In comparison to commonly used ranking loss and triplet loss for image retrieval, the multi-label cross-entropy loss can be considered as a separately training classifier for each tag, which is more robust to handle such annotation noise, and thus achieves better results.
\begin{figure}[!t]
\centering
\includegraphics[width=3.3in]{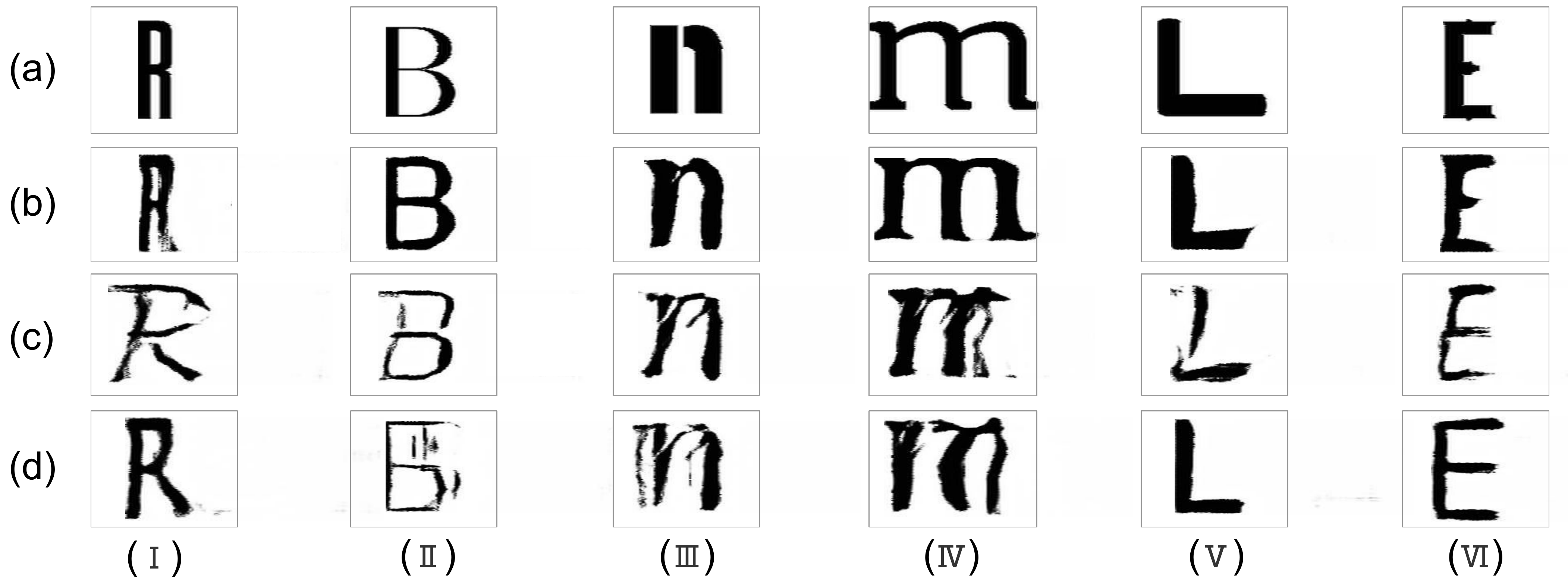}
\caption{Comparison of the reconstructed glyph images using different feature nodes: (a) ground-truth (input) glyph, (b) reconstructed glyph from the top-100 feature nodes of the input's own attention map, (c) reconstructed glyph from the bottom-100 feature nodes of the input's own attention map, (d) reconstructed glyph from the top-100 feature nodes of the adjacent glyph's attention map (``\uppercase\expandafter{\romannumeral1}'' $\leftrightarrow$ ``\uppercase\expandafter{\romannumeral2}'', ``\uppercase\expandafter{\romannumeral3}'' $\leftrightarrow$ ``\uppercase\expandafter{\romannumeral4}'', ``\uppercase\expandafter{\romannumeral5}'' $\leftrightarrow$ ``\uppercase\expandafter{\romannumeral6}''). Each column represents an input/reconstructed group.  }
\label{fig:stage2att}
\vspace{-3mm}
\end{figure}

The bottom part of Table~\ref{tab:result-1} shows the ablation study of our method. With individual modules added one by one, our method gets steady improvements, which demonstrates the effectiveness of each component. The generative feature learning module and the attentive feature selection module capture font specific features and thus learn a better font feature representation, which leads to improved retrieval results. The retrieval model handles the imbalanced tag frequency in multi-tag queries, so that the top ranked fonts can match all the tags in the query. We show some font retrieval results based on our full model in Figure~\ref{fig:example}. 


We show some qualitative results to further demonstrate the effectiveness of the proposed modules. We first verify that the attentive selection is effective in guiding the model to select useful and discriminated features for different fonts. Because the implicit attention mechanism is implemented at the node level, it is difficult to visualize as explicit attention mechanism. Therefore, we design a novel way for visualization. First, we train a GAN in the same way as Section~\ref{sec:collect} to reconstruct a glyph image from its extracted hidden feature. Given an input glyph image, we further restrict the GAN to reconstruct the glyph image by only using its top-100/bottom-100 hidden feature nodes with the highest/lowest attention weights from this attention map. We manually set the rest nodes to 0. For the last row in Figure~\ref{fig:stage2att}, we reconstruct each glyph image using its 100 feature nodes correspond to the highest attention weights on the attention map of another glyph image. From Figure~\ref{fig:stage2att}, we find that a glyph image reconstructed by the top-100 nodes based on its own attention map is more similar to the original input image. This serves our motivation for attentive feature selection, which is to adaptively select features that have effect on the input glyph image's tags and filter out the useless ones that cause error. In addition, the cross-attention generation results in the last row indicate that the model pays attention to different features for different fonts, instead of simply focusing on a uniform set of features.

\begin{figure}[!t]
\hspace{-1mm}\includegraphics[width=3.4in]{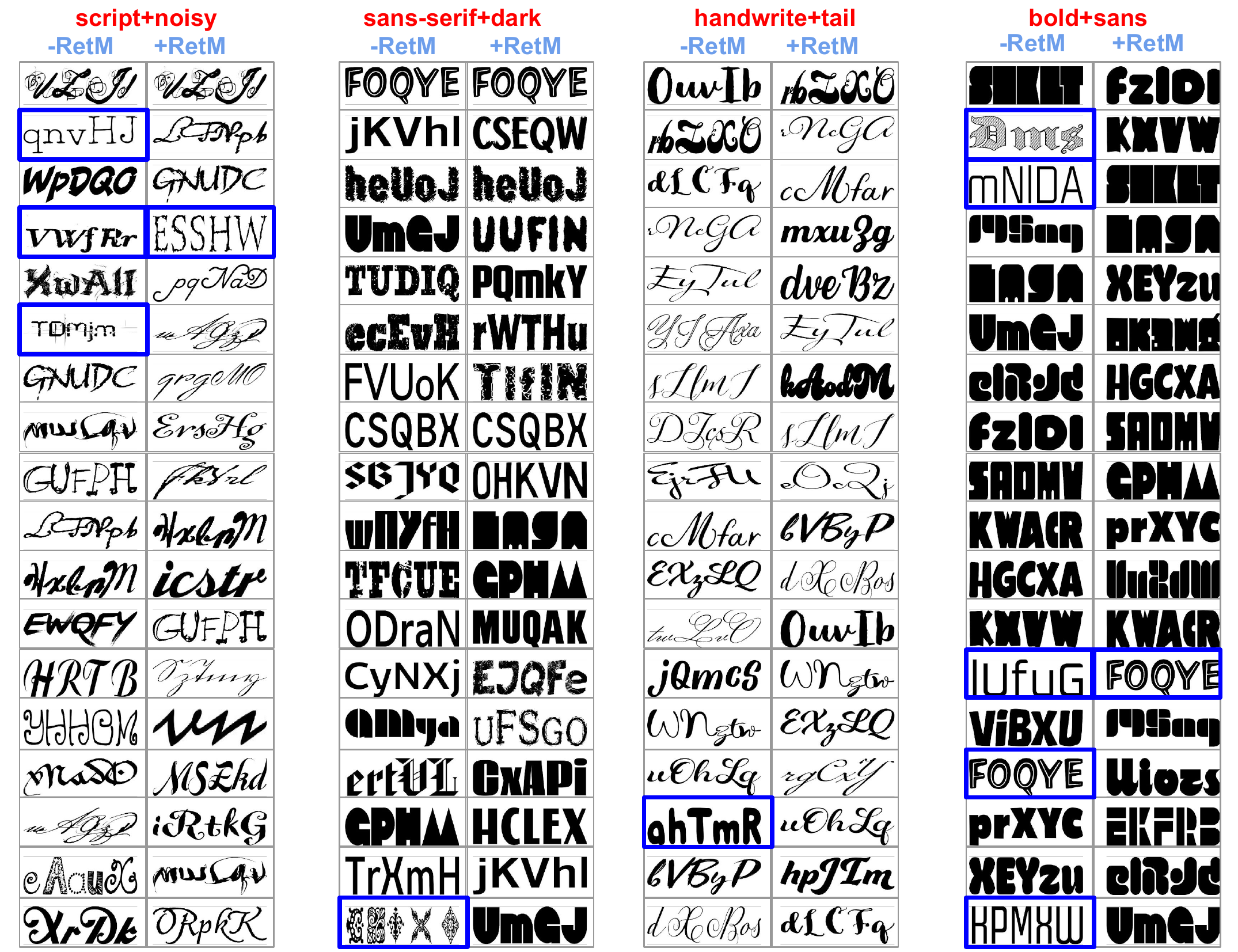}
\caption{Comparison of the models with (``+RetM'') and without (``-RetM'') the retrieval model on typical queries. The obvious failure cases are outlined by blue boxes.}
\label{fig:retexample}
\vspace{-1mm}
\end{figure}


Finally, Figure~\ref{fig:retexample} shows the different multi-tag retrieval results obtained with and without the retrieval model. Given a multi-tag query including a frequent tag ``script'' and an infrequent tag ``noisy'', our retrieval model can retrieve fonts which match both tags. While without the retrieval model, our retrieval results are less satisfactory, it recommends some fonts that don't have attribute ``script'' because of the tag dominance problem.

\subsection{AMT Test}

\begin{table}[tbp]
  \centering
  \begin{tabular}{|l|c|c|}
    \cline{1-3}
    Methods &Accuracy &Ave. Rank \\ \cline{1-3}
    Ours-Basic &44.49&1.81\\ \cline{1-3}
    \textbf{Ours} &\textbf{47.50}&\textbf{1.75}\\ \cline{1-3}
    \end{tabular}
\vspace{1mm}
\caption{\label{tab:result2} Comparison of our basic tag recognition model and full model on the AMT-test set.}
\vspace{-3mm}
\end{table}%

\begin{figure}[!t]
\centering
\hspace{-1mm}\includegraphics[width=3.3in]{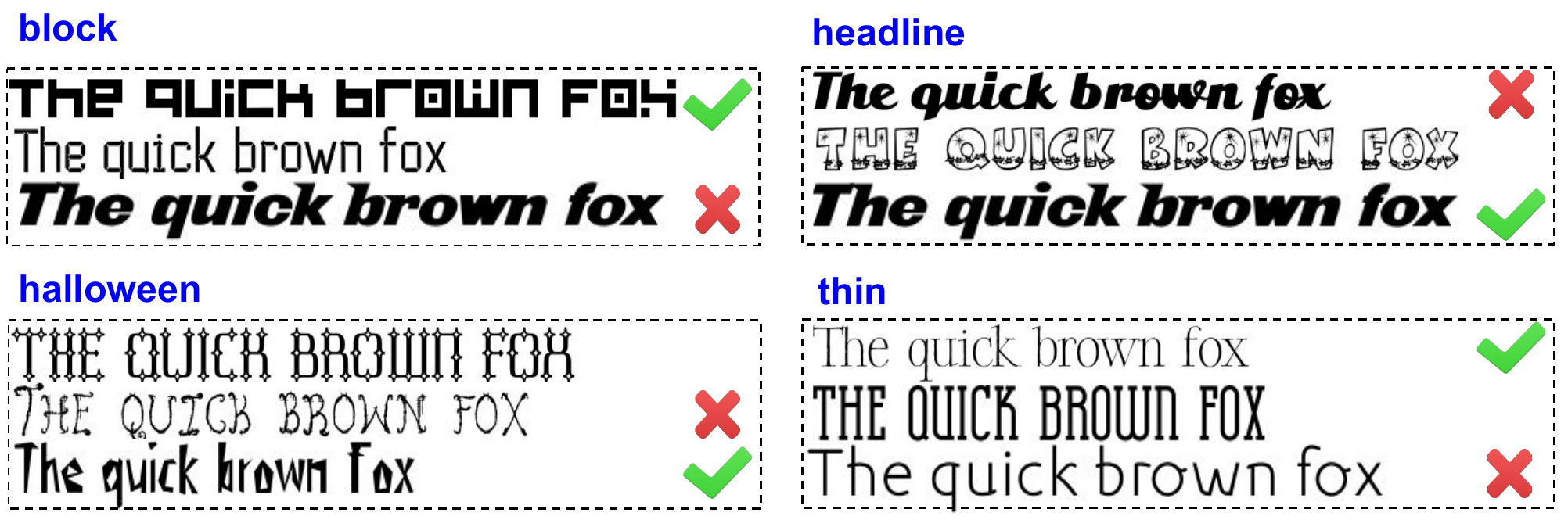}
\caption{Qualitative comparison of our basic recognition model and full model on typical groups of AMT-test. The fonts with ``\xmark" are falsely predicted by the basic recognition model, the ground-truth fonts with ``\cmark" are correctly predicted by the full model.}
\label{fig:amtexample}
\vspace{-3mm}
\end{figure}

Table~\ref{tab:result-1} shows that our basic recognition model is a strong baseline and much better than previous retrieval methods. We only compare our full model with the basic recognition model on the AMT-test set. Overall, the task on AMT-test is difficult because all fonts in the group of a tag is originally labeled with the tag. The model needs to select the fonts that are more relevant. As shown in Table~\ref{tab:result2}, our full model  achieves better performance than the basic version. It indicates that for the top-related fonts toward one tag, the ranking of the full model is still more consistent with the human judgment. Some qualitative results are illustrated in Figure~\ref{fig:amtexample}.

%

\section{Conclusion}

In this paper, we study the problem of tag-based font retrieval. We collect a large-scale font tagging dataset. We propose a joint recognition-retrieval model. In particular, a recognition model is first trained to predict the tag probabilities of a font. With the guidance of the generative feature learning and attentive feature selection mechanisms, the model adaptively selects the general and significant information of the font and makes  better tag probability prediction. A retrieval model is further integrated to map  tag probabilities to font-query affinity scores. Extensive qualitative and quantitative evaluations validate the  effectiveness of our model for font retrieval.

\section{Acknowledgment}
This work is partially supported by NSF award 
\#1704337.

{\small
\bibliographystyle{ieee_fullname}
\balance
\bibliography{egbib}

\begin{thebibliography}{10}\itemsep=-1pt

\bibitem{anderson2017bottom}
Peter Anderson, Xiaodong He, Chris Buehler, Damien Teney, Mark Johnson, Stephen
  Gould, and Lei Zhang.
\newblock Bottom-up and top-down attention for image captioning and vqa.
\newblock {\em arXiv preprint arXiv:1707.07998}, 2017.

\bibitem{azadi2018multi}
Samaneh Azadi, Matthew Fisher, Vladimir Kim, Zhaowen Wang, Eli Shechtman, and
  Trevor Darrell.
\newblock Multi-content gan for few-shot font style transfer.
\newblock In {\em Proceedings of the IEEE Conference on Computer Vision and
  Pattern Recognition}, volume~11, page~13, 2018.

\bibitem{chen2014large}
Guang Chen, Jianchao Yang, Hailin Jin, Jonathan Brandt, Eli Shechtman, Aseem
  Agarwala, and Tony~X Han.
\newblock Large-scale visual font recognition.
\newblock In {\em Proceedings of the IEEE Conference on Computer Vision and
  Pattern Recognition}, pages 3598--3605, 2014.

\bibitem{friedman2001greedy}
Jerome~H Friedman.
\newblock Greedy function approximation: a gradient boosting machine.
\newblock {\em Annals of statistics}, pages 1189--1232, 2001.

\bibitem{frome2013devise}
Andrea Frome, Greg~S Corrado, Jon Shlens, Samy Bengio, Jeff Dean, Tomas
  Mikolov, et~al.
\newblock Devise: A deep visual-semantic embedding model.
\newblock In {\em Advances in neural information processing systems}, pages
  2121--2129, 2013.

\bibitem{goodfellow2014generative}
Ian Goodfellow, Jean Pouget-Abadie, Mehdi Mirza, Bing Xu, David Warde-Farley,
  Sherjil Ozair, Aaron Courville, and Yoshua Bengio.
\newblock Generative adversarial nets.
\newblock In {\em Advances in neural information processing systems}, pages
  2672--2680, 2014.

\bibitem{gu2018look}
Jiuxiang Gu, Jianfei Cai, Shafiq Joty, Li Niu, and Gang Wang.
\newblock Look, imagine and match: Improving textual-visual cross-modal
  retrieval with generative models.
\newblock In {\em Proceedings of the IEEE Conference on Computer Vision and
  Pattern Recognition}, pages 7181--7189, 2018.

\bibitem{he2016deep}
Kaiming He, Xiangyu Zhang, Shaoqing Ren, and Jian Sun.
\newblock Deep residual learning for image recognition.
\newblock In {\em Proceedings of the IEEE conference on computer vision and
  pattern recognition}, pages 770--778, 2016.

\bibitem{isola2017image}
Phillip Isola, Jun-Yan Zhu, Tinghui Zhou, and Alexei~A Efros.
\newblock Image-to-image translation with conditional adversarial networks.
\newblock {\em arXiv preprint}, 2017.

\bibitem{kim2016multimodal}
Jin-Hwa Kim, Sang-Woo Lee, Donghyun Kwak, Min-Oh Heo, Jeonghee Kim, Jung-Woo
  Ha, and Byoung-Tak Zhang.
\newblock Multimodal residual learning for visual qa.
\newblock In {\em Advances in Neural Information Processing Systems}, pages
  361--369, 2016.

\bibitem{imagenet12}
Alex Krizhevsky, Ilya Sutskever, and Geoffrey~E. Hinton.
\newblock Imagenet classification with deep convolutional neural networks.
\newblock In {\em Advances in neural information processing systems}, 2012.

\bibitem{lampert2009learning}
Christoph~H Lampert, Hannes Nickisch, and Stefan Harmeling.
\newblock Learning to detect unseen object classes by between-class attribute
  transfer.
\newblock In {\em Computer Vision and Pattern Recognition, 2009. CVPR 2009.
  IEEE Conference on}, pages 951--958. IEEE, 2009.

\bibitem{lee2018stacked}
Kuang-Huei Lee, Xi Chen, Gang Hua, Houdong Hu, and Xiaodong He.
\newblock Stacked cross attention for image-text matching.
\newblock {\em arXiv preprint arXiv:1803.08024}, 2018.

\bibitem{li2017identity}
Shuang Li, Tong Xiao, Hongsheng Li, Wei Yang, and Xiaogang Wang.
\newblock Identity-aware textual-visual matching with latent co-attention.
\newblock In {\em Computer Vision (ICCV), 2017 IEEE International Conference
  on}, pages 1908--1917. IEEE, 2017.

\bibitem{li2017person}
Shuang Li, Tong Xiao, Hongsheng Li, Bolei Zhou, Dayu Yue, and Xiaogang Wang.
\newblock Person search with natural language description.
\newblock {\em arXiv preprint arXiv:1702.05729}, 2017.

\bibitem{liu2018font}
Yang Liu, Zhaowen Wang, Hailin Jin, and Ian Wassell.
\newblock Multi-task adversarial network for disentangled feature learning.
\newblock In {\em CVPR}, 2018.

\bibitem{Liu_2018_ECCV}
Yang Liu, Zhaowen Wang, Hailin Jin, and Ian Wassell.
\newblock Synthetically supervised feature learning for scene text recognition.
\newblock In {\em The European Conference on Computer Vision (ECCV)}, September
  2018.

\bibitem{lu2017knowing}
Jiasen Lu, Caiming Xiong, Devi Parikh, and Richard Socher.
\newblock Knowing when to look: Adaptive attention via a visual sentinel for
  image captioning.
\newblock In {\em Proceedings of the IEEE Conference on Computer Vision and
  Pattern Recognition (CVPR)}, volume~6, page~2, 2017.

\bibitem{ma2018gan}
Shuang Ma, Jianlong Fu, Chang~Wen Chen, and Tao Mei.
\newblock Da-gan: Instance-level image translation by deep attention generative
  adversarial networks.
\newblock In {\em Proceedings of the IEEE Conference on Computer Vision and
  Pattern Recognition}, pages 5657--5666, 2018.

\bibitem{mirza2014conditional}
Mehdi Mirza and Simon Osindero.
\newblock Conditional generative adversarial nets.
\newblock {\em arXiv preprint arXiv:1411.1784}, 2014.

\bibitem{o2014exploratory}
Peter O'Donovan, J{\=a}nis L{\=\i}beks, Aseem Agarwala, and Aaron Hertzmann.
\newblock Exploratory font selection using crowdsourced attributes.
\newblock {\em ACM Transactions on Graphics (TOG)}, 33(4):92, 2014.

\bibitem{reed2016learning}
Scott Reed, Zeynep Akata, Honglak Lee, and Bernt Schiele.
\newblock Learning deep representations of fine-grained visual descriptions.
\newblock In {\em Proceedings of the IEEE Conference on Computer Vision and
  Pattern Recognition}, pages 49--58, 2016.

\bibitem{socher2013zero}
Richard Socher, Milind Ganjoo, Christopher~D Manning, and Andrew Ng.
\newblock Zero-shot learning through cross-modal transfer.
\newblock In {\em Advances in neural information processing systems}, pages
  935--943, 2013.

\bibitem{tran2017disentangled}
Luan Tran, Xi Yin, and Xiaoming Liu.
\newblock Disentangled representation learning gan for pose-invariant face
  recognition.
\newblock In {\em Proceedings of the IEEE Conference on Computer Vision and
  Pattern Recognition}, pages 1415--1424, 2017.

\bibitem{WahCUB_200_2011}
C. Wah, S. Branson, P. Welinder, P. Perona, and S. Belongie.
\newblock {The Caltech-UCSD Birds-200-2011 Dataset}.
\newblock Technical Report CNS-TR-2011-001, California Institute of Technology,
  2011.

\bibitem{wang2017adversarial}
Bokun Wang, Yang Yang, Xing Xu, Alan Hanjalic, and Heng~Tao Shen.
\newblock Adversarial cross-modal retrieval.
\newblock In {\em Proceedings of the 2017 ACM on Multimedia Conference}, pages
  154--162. ACM, 2017.

\bibitem{wang2015deepfont}
Zhangyang Wang, Jianchao Yang, Hailin Jin, Eli Shechtman, Aseem Agarwala,
  Jonathan Brandt, and Thomas~S Huang.
\newblock Deepfont: Identify your font from an image.
\newblock In {\em Proceedings of the 23rd ACM international conference on
  Multimedia}, pages 451--459. ACM, 2015.

\bibitem{xian2017zero}
Yongqin Xian, Bernt Schiele, and Zeynep Akata.
\newblock Zero-shot learning-the good, the bad and the ugly.
\newblock {\em arXiv preprint arXiv:1703.04394}, 2017.

\bibitem{yang2018learning}
Flood Sung~Yongxin Yang, Li Zhang, Tao Xiang, Philip~HS Torr, and Timothy~M
  Hospedales.
\newblock Learning to compare: Relation network for few-shot learning.
\newblock 2018.

\bibitem{yang2016stacked}
Zichao Yang, Xiaodong He, Jianfeng Gao, Li Deng, and Alex Smola.
\newblock Stacked attention networks for image question answering.
\newblock In {\em Proceedings of the IEEE Conference on Computer Vision and
  Pattern Recognition}, pages 21--29, 2016.

\bibitem{you2016image}
Quanzeng You, Hailin Jin, Zhaowen Wang, Chen Fang, and Jiebo Luo.
\newblock Image captioning with semantic attention.
\newblock In {\em Proceedings of the IEEE conference on computer vision and
  pattern recognition}, pages 4651--4659, 2016.

\bibitem{zhang2017learning}
Li Zhang, Tao Xiang, Shaogang Gong, et~al.
\newblock Learning a deep embedding model for zero-shot learning.
\newblock 2017.

\bibitem{zhang2018attention}
Xi Zhang, Hanjiang Lai, and Jiashi Feng.
\newblock Attention-aware deep adversarial hashing for cross-modal retrieval.
\newblock In {\em European Conference on Computer Vision}, pages 614--629.
  Springer, 2018.

\bibitem{zhang2018separating}
Yexun Zhang, Ya Zhang, and Wenbin Cai.
\newblock Separating style and content for generalized style transfer.
\newblock In {\em Proceedings of the IEEE Conference on Computer Vision and
  Pattern Recognition}, volume~1, 2018.

\bibitem{zheng2017dual}
Zhedong Zheng, Liang Zheng, Michael Garrett, Yi Yang, and Yi-Dong Shen.
\newblock Dual-path convolutional image-text embedding.
\newblock {\em arXiv preprint arXiv:1711.05535}, 2017.

\bibitem{zhou2018places}
Bolei Zhou, Agata Lapedriza, Aditya Khosla, Aude Oliva, and Antonio Torralba.
\newblock Places: A 10 million image database for scene recognition.
\newblock {\em IEEE transactions on pattern analysis and machine intelligence},
  40(6):1452--1464, 2018.

\bibitem{zhu2017unpaired}
Jun-Yan Zhu, Taesung Park, Phillip Isola, and Alexei~A Efros.
\newblock Unpaired image-to-image translation using cycle-consistent
  adversarial networks.
\newblock {\em arXiv preprint}, 2017.

\bibitem{zhu2001font}
Yong Zhu, Tieniu Tan, and Yunhong Wang.
\newblock Font recognition based on global texture analysis.
\newblock {\em IEEE Transactions on pattern analysis and machine intelligence},
  23(10):1192--1200, 2001.

\bibitem{zramdini1998optical}
Abdelwahab Zramdini and Rolf Ingold.
\newblock Optical font recognition using typographical features.
\newblock {\em IEEE Transactions on Pattern Analysis \& Machine Intelligence},
  (8):877--882, 1998.

\end{thebibliography}
}

\clearpage
\section{Supplementary Material}

In this document, we provide additional materials to supplement our paper ``Large-scale Tag-based Font Retrieval with Generative Feature Learning''. In the first section, we provide more details about our collected font retrieval dataset. In the second section, we describe the evaluation measurement of Myfonts-test set. In the third section, we provide the weight initialization details of the attention module and the retrieval model. In the fourth section, we comprehensively illustrate the retrieval performance of the proposed model on typical single-tag and multi-tag queries. 

\subsection{Dataset Supplementary Information}
In Section 3 of our paper, we present a large-scale tag-based font retrieval dataset, which is collected from MyFonts. After tag preprocessing, the dataset finally contains 1824 tags for font description. Table~\ref{tab:dataset} provides us deeper insight into these tags by showing the top-200 frequent tags. It can be seen that the dataset contains meaningful tags that cover different aspects of a font, such as its category (e.g ``sans-serif'', ``script'', ``handwrite''), appearance (e.g. ``heavy'',``outline'',``round''), utility (e.g. ``poster'', ``magazine'', ``logo'') and other special features (e.g. ``kid'', ``romantic'', ``cartoon'').

\begin{figure*}[!t]

\centering
\includegraphics[width=5.0in]{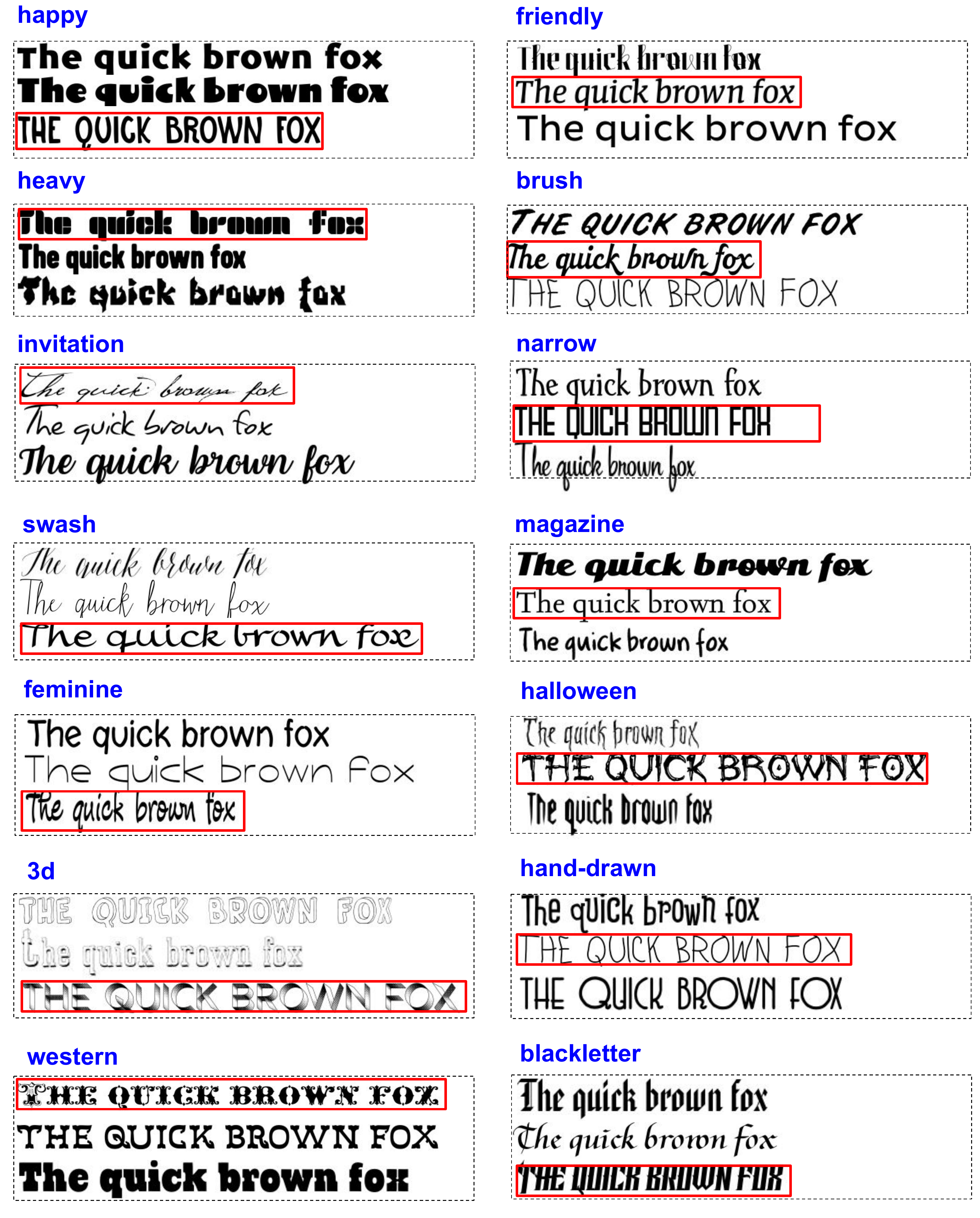}
\caption{Group examples of the collected tagging set. For each group of a tag, the ground-truth font is in the red box.}
\label{fig:tagging}
\vspace{-4mm}
\end{figure*}

On the other hand, we collect a tagging set with ranking information via Amazon Mechanical Turk (AMT) as a complement for evaluation. The detailed process of collecting the tagging set is described in the main submission. Roughly speaking, this set contains 1661 groups. Each group includes three fonts related to a specific tag, and is finally labeled a most matching one agreed by all the workers. We show a large number of group examples in Figure~\ref{fig:tagging} to present the tagging set in detail.

\subsection{Measurement of MyFonts-test Set}
We evaluate different models' performance on the MyFonts-test set by two standard measures, mean average precision (mAP) and Normalized Discounted cumulative gain (nDCG). For average precision (AP), given a query $q$, assuming that the total $H$ positive fonts \{$f_{1}$, $f_{2}$,..., $f_{H}$\} in the test set have affinity score ranks \{$r_{1}$, $r_{2}$,..., $r_{H}$\}, the average precision score of $q$ ($AP_{q}$) is computed as: $AP_{q} = \frac{1}{H}\sum_{h=1}^{H}{\frac{h}{r_{h}}}$. For nDCG, given the font relevance \{$rel_{1}$, $rel_{2}$,..., $rel_{S}$\} for the total $S$ test fonts, which have affinity score ranks $\{1, 2, ..., S\}$ on query $q$, $nDCG_{q}$ is computed as: $DCG_{q} = \sum_{p=1}^{S}\frac{2^{rel_{p}}-1}{\log_{2}(p+1)}$, $nDCG_{q} = \frac{DCG_{q}}{IDCG_{q}}$, where $IDCG_{q}$ is the maximum possible value of $DCG_{q}$ for different ranking results on $q$. The font relevance for a positive font is set to 1, for a negative font, it is set to 0.

In our experiments, given a set of test queries, we compute the mean value of AP (mAP) and nDCG for all queries as the final mAP and nDCG scores.

\subsection{Weight Initialization}
In the training process, we find that the weight initialization of the attention module and the retrieval model can make effect on the final performance. For the attention module that contains a fully-connected layer followed by a sigmoid unit, the weights of the fully-connected layer are initialized using a normal distribution $(\mu, \sigma)$ where $\mu = 0$ and $\sigma = 5$. The retrieval model contains two fully-connected layers whose dimensions are set as $N$ and $1$. $N$ is the total tag vocabulary size. The first layer with a ReLU unit maps the $N$-dimensional query-based tag probability vector to a $N$-dimensional feature vector. The second layer with a sigmoid unit then maps it to the final affinity score. We use a $N \times N$ identity matrix to initialize the weights of the first layer, and use a normal distribution with $\mu = 1$ and $\sigma = 0.02$ to initialize the weights of the second layer.

\subsection{Additional qualitative results}

In this section, we illustrate a great number of font retrieval results of the proposed model as the supplement of Figure 6 in the main submission. We test the model's performance on typical single-tag and multi-tag queries, the top-20 retrieved fonts for each query are shown as Figure~\ref{fig:example1}. These results demonstrate the effectiveness of our model to retrieve the top corresponding fonts for a wide range of tags.

\begin{table*}[htbp]
  \small\centering
  \vspace{-4mm}
  \caption{\label{tab:dataset} Top-200 frequent tags for the collected dataset.}
  
  \begin{threeparttable}
  \scalebox{0.8}{
  \begin{tabular}{|c|c|c|c|c|c|c|c|c|c|}
    \cline{1-10}
    decorative&display&headline&poster&sans-serif&magazine&modern&legible&retro&script\\ \cline{1-10}
elegant&informal&serif&handwrite&logo&geometric&funny&vintage&contemporary&clean\\ \cline{1-10}
bold&sans&alternate&package&sketch&brand&heavy&text&round&invitation\\ \cline{1-10}
ligature&letter&advertise&fun&friendly&calligraphy&hand&rough&wed&swash\\ \cline{1-10}
ornament&casual&cool&brush&handmade&fashion&calligraphic&commercial&narrow&cursive\\ \cline{1-10}
book&signage&comic&grunge&fancy&art-deco&hand-drawn&editorial&corporate&fashionable\\ \cline{1-10}
cute&condense&kid&organic&multilingual&feminine&monoline&slab-serif&connect&antique\\ \cline{1-10}
cyrillic&pen&logotype&title&all-caps&german&news&ink&square&symbol\\ \cline{1-10}
playful&formal&grotesk&soft&futuristic&child&humanist&thin&stylish&classic\\ \cline{1-10}
scrapbook&technical&light&black&wide&simple&techno&ancient&food&unique\\ \cline{1-10}
stencil&american&design&grotesque&dingbat&opentype&1930s&unusual&block&picture\\ \cline{1-10}
cartoon&italic&small-caps&1800s&outline&computer&music&illustration&capital-sharp-s&web\\ \cline{1-10}
versal-eszett&1950s&masculine&valentine&party&lively&sanserif&creative&irregular&romantic\\ \cline{1-10}
curly&wood-type&blackletter&1960s&happy&1920s&label&greet&newspaper&gothic\\ \cline{1-10}
print&distress&sharp&ornamental&fresh&sport&workhorse&delicate&capital&1940s\\ \cline{1-10}
screen&cap&natural&brush-drawn&love&sign-painting&inline&hipster&engrave&menu\\ \cline{1-10}
fat&art-nouveau&experimental&oldstyle&industrial&neutral&1970s&bouncy&crazy&wild\\ \cline{1-10}
sign&caps-only&flourish&versatile&minimal&movie&french&linear&urban&roman\\ \cline{1-10}
slab&film&publish&angular&modular&beautiful&art&expressive&texture&deco\\ \cline{1-10}
graceful&mechanical&letterpress&paint&western&wild-west&dynamic&game&greek&draw\\ \cline{1-10}
    \end{tabular}}%
  \end{threeparttable}

\end{table*}%

\begin{figure*}[!b]
\vspace{-4mm}
\centering
\includegraphics[width=6.6in]{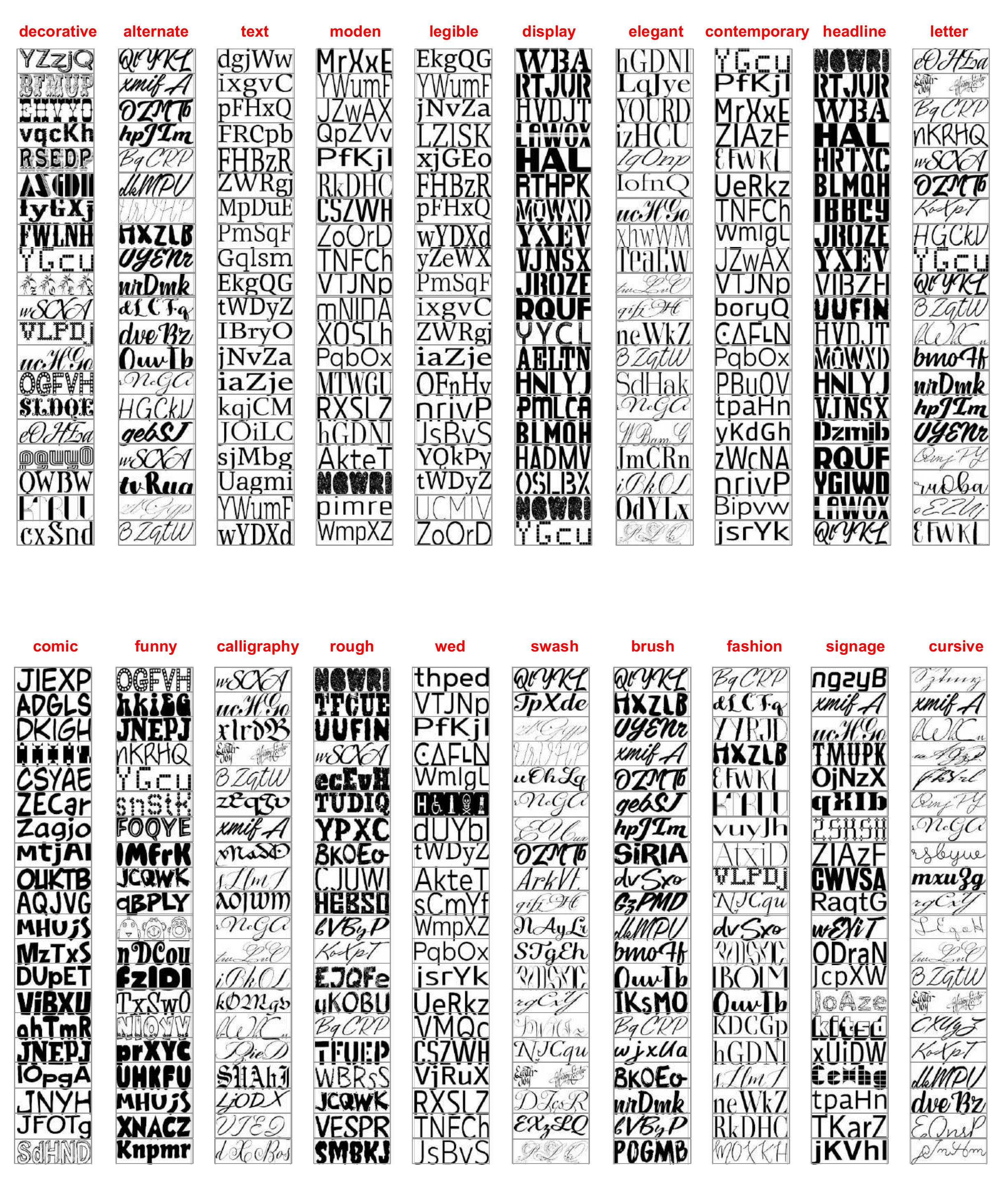}
\label{fig:example}
\vspace{-4mm}
\end{figure*}
\begin{figure*}[!t]
\vspace{-4mm}
\centering
\includegraphics[width=6.6in]{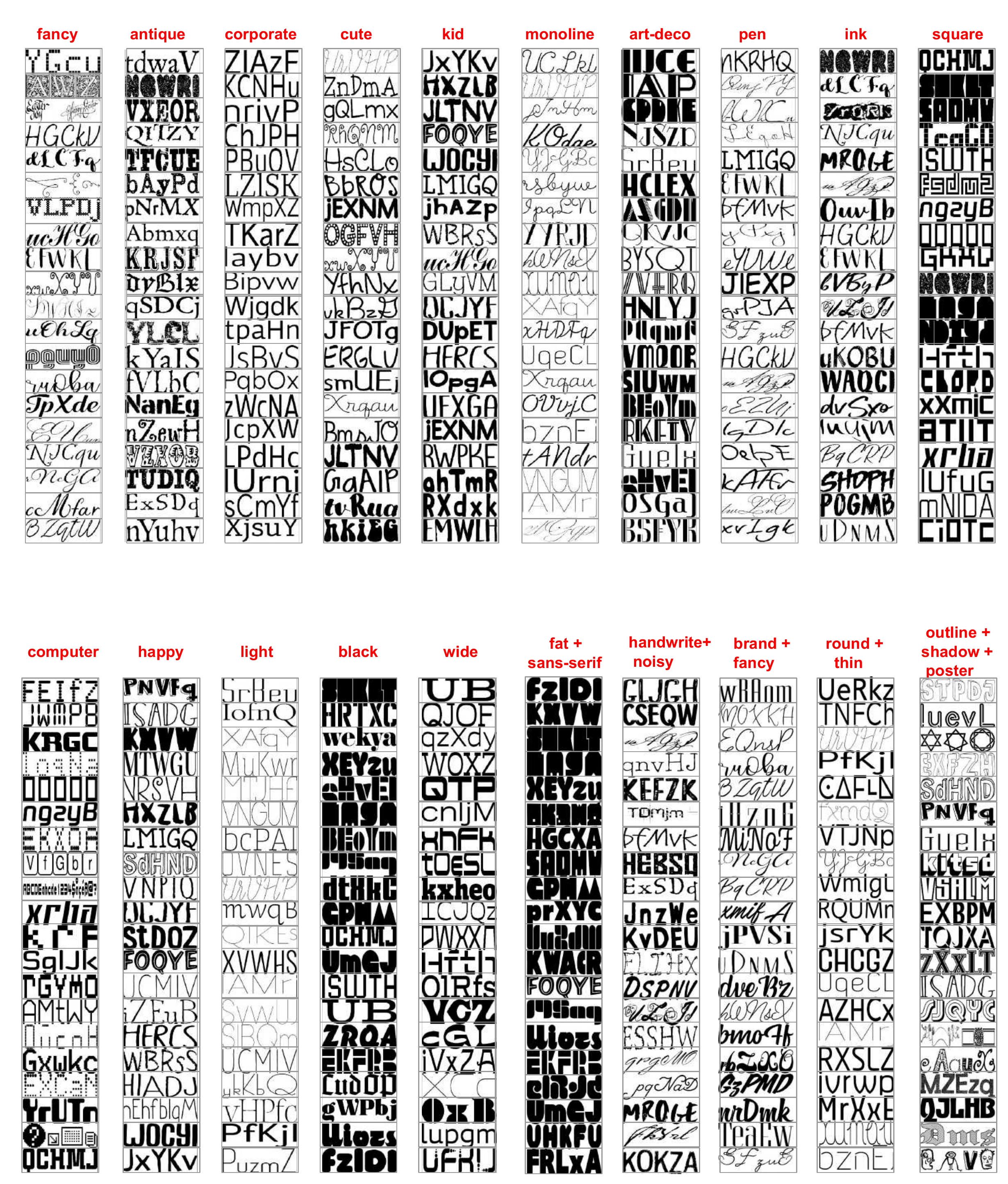}
\caption{Font retrieval results of the proposed model on typical single-tag and multi-tag queries.}
\label{fig:example1}
\vspace{-4mm}
\end{figure*}

\end{document}